\documentclass[11pt,a4paper,draft]{article}
\usepackage[hyperref]{naaclhlt2018}
\usepackage{times}
\usepackage{latexsym}

\usepackage[utf8]{inputenc}
\usepackage[vietnamese,main=english]{babel}
\usepackage{url}
\usepackage{amsmath}
\usepackage{amsfonts}
\usepackage{multirow}
\usepackage{color,colortbl}
\usepackage{booktabs}
\usepackage[draft]{graphicx}

\newcommand{\VI}{\foreignlanguage{vietnamese}}

\newcommand{\NOTE}[1]{}

\aclfinalcopy 

\title{Introducing two Vietnamese Datasets for Evaluating Semantic Models of (Dis-)Similarity and Relatedness}

\author{Kim Anh Nguyen \and Sabine Schulte im Walde \and Ngoc Thang Vu \\
	    Institut f\"ur Maschinelle Sprachverarbeitung\\
	    Universit\"at Stuttgart\\
	    Pfaffenwaldring 5B, 70569 Stuttgart, Germany\\
	    {\{\tt nguyenkh,schulte,thangvu\}@ims.uni-stuttgart.de}}

\date{}

\begin{document}
\maketitle
\newenvironment{vietnamese}
  {\begin{otherlanguage}{vietnamese}}
  {\end{otherlanguage}}
\newenvironment{vietnamese*}
  {\begin{otherlanguage*}{vietnamese}}
  {\end{otherlanguage*}}

\begin{abstract}
  We present two novel datasets for the low-resource language
  Vietnamese to assess models of semantic similarity: \textit{ViCon}
  comprises pairs of synonyms and antonyms across word classes, thus
  offering data to distinguish between similarity and
  dissimilarity. \textit{ViSim-400} provides degrees of similarity
  across five semantic relations, as rated by human judges. The two
  datasets are verified through standard co-occurrence and neural
  network models, showing results comparable to the respective English
  datasets.
\end{abstract}

\vspace{+3mm}
\section{Introduction}
\label{introduction}

Computational models that distinguish between semantic similarity and
semantic relatedness \citep{Budanitsky/Hirst:06} are important for
many NLP applications, such as the automatic generation of
dictionaries, thesauri, and
ontologies~\cite{Biemann:05,Cimiano:05,Li:06}, and machine
translation~\cite{He:08,Marton:09}. In order to evaluate these models,
gold standard resources with word pairs have to be collected
(typically across semantic relations such as synonymy, hypernymy,
antonymy, co-hyponymy, meronomy, etc.) and annotated for their degree
of similarity via human judgements.

The most prominent examples of gold standard similarity resources for
English are the Rubenstein \& Goodenough (RG) dataset
\cite{Rubenstein/Goodenough:1965}, the TOEFL test questions
\cite{Landauer/Dutnais:1997}, WordSim-353 \cite{Finkelstein:2001}, MEN
\cite{Bruni:2012}, SimLex-999 \cite{Hill:2015}, and the lexical
contrast datasets by \cite{NguyenEtAl:2016,NguyenEtAl:2017}. For other
languages, resource examples are the translation of the RG dataset to
German \cite{Gurevych:2005}, the German dataset of paradigmatic
relations \cite{Scheible/Schulteimwalde:2014}, and the translation of
WordSim-353 and SimLex-999 to German, Italian and Russian
\cite{Leviant/Reichart:2015}. However, for low-resource languages
there is still a lack of such datasets, which we aim to fill for
Vietnamese, a language without morphological marking such as case,
gender, number, and tense, thus differing strongly from Western
European languages.

We introduce two novel datasets for Vietnamese: a dataset of lexical
contrast pairs \textbf{ViCon} to distinguish between similarity
(synonymy) and dissimilarity (antonymy), and a dataset of semantic
relation pairs \textbf{ViSim-400} to reflect the continuum between
similarity and relatedness. The two datasets are publicly
available.\footnote{\scriptsize
  \url{www.ims.uni-stuttgart.de/data/vnese_sem_datasets}} Moreover, we
verify our novel datasets through standard and neural co-occurrence
models, in order to show that we obtain a similar behaviour as for the
corresponding English datasets \textit{SimLex-999}~\cite{Hill:2015},
and the lexical contrast dataset (henceforth \textit{LexCon}),
cf.~\newcite{NguyenEtAl:2016}.

\vspace{+3mm}
\section{Related Work}
\label{related-work}

Over the years a number of datasets have been collected for studying
and evaluating semantic similarity and semantic relatedness. For
English, Rubenstein and
Goodenough~\shortcite{Rubenstein/Goodenough:1965} presented a small
dataset (RG) of 65 noun pairs. For each pair, the degree of similarity
in meaning was provided by 15 raters. The RG dataset is assumed to
reflect similarity rather than relatedness. Finkelstein et
al. \shortcite{Finkelstein:2001} created a set of 353 English
noun-noun pairs (WordSim-353)\footnote{\scriptsize
  \url{www.cs.technion.ac.il/\~gabr/resources/data/wordsim353}}, where
each pair was rated by 16 subjects according to the degree of semantic
relatedness on a scale from 0 to 10. \newcite{Bruni:2012} introduced a
large test collection called MEN\footnote{\scriptsize
  \url{clic.cimec.unitn.it/\~elia.bruni/MEN}}. Similar to WordSim-353,
the authors refer to both similarity and relatedness when describing
the MEN dataset, although the annotators were asked to rate the pairs
according to relatedness. Unlikely the construction of the RG and
WordSim-353 datasets, each pair in the MEN dataset was only evaluated
by one rater who ranked it for relatedness relative to 50 other pairs
in the dataset. Recently, \newcite{Hill:2015} presented SimLex-999, a
gold standard resource for the evaluation of semantic representations
containing similarity ratings of word pairs across different
part-of-speech categories and concreteness levels. The construction of
SimLex-999 was motivated by two factors, (i) to consistently quantify
similarity, as distinct from association, and apply it to various
concept types, based on minimal intuitive instructions, and (ii) to
have room for the improvement of state-of-the-art models which had
reached or surpassed the human agreement ceiling on WordSim-353 and
MEN, the most popular existing gold standards, as well as on
RG. \newcite{Scheible/Schulteimwalde:2014} presented a collection of
semantically related word pairs for German and
English,\footnote{\scriptsize
  \url{www.ims.uni-stuttgart.de/data/sem-rel-database/}} which was
compiled via Amazon Mechanical Turk
(AMT)\footnote{\scriptsize{\url{www.mturk.com}}} human judgement
experiments and comprises (i) a selection of targets across word
classes balanced for semantic category, polysemy, and corpus
frequency, (ii) a set of human-generated semantically related word
pairs (synonyms, antonyms, hypernyms) based on the target units, and
(iii) a subset of the generated word pairs rated for their relation
strength, including positive and negative relation evidence.

For other languages, only a few gold standard sets with scored word
pairs exist. Among others, \newcite{Gurevych:2005} replicated
Rubenstein and Goodenough’s experiments after translating the original
65 word pairs into German. In later work, \newcite{Gurevych:2006} used
the same experimental setup to increase the number of word pairs to
350. \newcite{Leviant/Reichart:2015} translated two prominent
evaluation sets, WordSim-353 (association) and SimLex-999 (similarity)
from English to Italian, German and Russian, and collected the scores
for each dataset from the respective native speakers via
crowdflower\footnote{\scriptsize \url{www.crowdflower.com/}}.

\section{Dataset Design}
\label{design}

\subsection{Criteria}
\label{criteria}

\textit{Semantic similarity} is a narrower concept than
\textit{semantic relatedness} and holds between lexical terms with
similar meanings. Strong similarity is typically observed for the
lexical relations of synonymy and co-hyponymy. For example, in
Vietnamese ``\VI{đội}" (\textit{team}) and ``\VI{nhóm}"
(\textit{group}) represents a synonym pair; ``\VI{ô\_tô}"
(\textit{car}) and ``\VI{xe\_đạp}" (\textit{bike}) is a co-hyponymy
pair.  More specifically, words in the pair ``\VI{ô\_tô}"
(\textit{car}) and ``\VI{xe\_đạp}" (\textit{bike}) share several
features such as physical (e.g. \VI{bánh\_xe} / \textit{wheels}) and
functional (e.g. \VI{vận\_tải} / \textit{transport}), so that the two
Vietnamese words are interchangeable regarding the kinds of
transportation. The concept of semantic relatedness is broader and
holds for relations such as meronymy, antonymy, functional
association, and other ``non-classical relations"
\cite{Morris/Hirst:2004}. For example, ``\VI{ô\_tô}" (\textit{car})
and ``\VI{xăng\_dầu}" (\textit{petrol}) represent a meronym pair. In
contrast to similarity, this meronym pair expresses a clearly
functional relationship; the words are strongly associated with each
other but not similar.

Empirical studies have shown that the predictions of distributional
models as well as humans are strongly related to the part-of-speech
(POS) category of the learned concepts. Among others,
\newcite{Gentner:06} showed that verb concepts are harder to learn by
children than noun concepts.

Distinguishing antonymy from synonymy is one of the most difficult
challenges. While antonymy represents words which are strongly
associated but highly dissimilar to each other, synonymy refers to
words that are highly similar in meaning. However, antonyms and
synonyms often occur in similar context, as they are interchangeable
in their substitution.

\subsection{Resource for Concept Choice:\\\hspace*{+9mm}Vietnamese Computational Lexicon}
\label{vcl}

The Vietnamese Computational Lexicon (VCL)\footnote{\scriptsize
  \url{https://vlsp.hpda.vn/demo/?page=vcl}} \cite{NguyenTMH:2006} is
a common linguistic database which is freely and easily exploitable
for automatic processing of the Vietnamese language. VCL contains
35,000 words corresponding to 41,700 concepts, accompanied by
morphological, syntactic and semantic information. The morphological
information consists of 8 morphemes including simple word, compound
word, reduplicative word, multi-word expression, loan word,
abbreviation, bound morpheme, and symbol. For example, ``\VI{bàn}"
(\textit{table}) is a simple word with definition ``\VI{đồ thường làm
  bằng gỗ, có mặt phẳng và chân đỡ \ldots}'' (\textit{pieces of wood,
  flat and supported by one or more legs \ldots}). The syntactic
information describes part-of-speech, collocations, and
subcategorisation frames. The semantic information includes two types
of constraints: logical and semantic. The logical constraint provides
category meaning, synonyms and antonyms. The semantic constraint
provides argument information and semantic roles. For example,
``\VI{yêu}'' (\textit{love}) is a verb with category meaning
``\textit{emotion}'' and antonym ``\VI{ghét}'' (\textit{hate}).

VCL is the largest linguistic database of its kind for Vietnamese, and
it encodes various types of morphological, syntactic and semantic
information, so it presents a suitable starting point for the choice
of lexical units for our purpose.

\vspace{+3mm}
\subsection{Choice of Concepts}
\label{choice-concepts}

\subsubsection{Concepts in ViCon}

The choice of related pairs in this dataset was drawn from VCL in the
following way. We extracted all antonym and synonym pairs according to
the three part-of-speech categories: noun, verb and adjective. We then
randomly selected 600 adjective pairs (300 antonymous pairs and 300
synonymous pairs), 400 noun pairs (200 antonymous pairs and 200
synonymous pairs), and 400 verb pairs (200 antonymous pairs and 200
synonymous pairs). In each part-of-speech category, we balanced for
the size of morphological classes in VCL, for both antonymous and
synonymous pairs.

\subsubsection{Concepts in ViSim-400}

The choice of related pairs in this dataset was drawn from both the
VLC and the \textit{Vietnamese WordNet}\footnote{\scriptsize
  \url{http://viet.wordnet.vn/wnms/}} (VWN),
cf. \newcite{NguyenPT:2016}. We extracted all pairs of the three
part-of-speech categories: noun, verb and adjective, according to five
semantic relations: synonymy, antonymy, hypernymy, co-hoponymy and
meronymy. We then sampled 400 pairs for the ViSim-400 dataset,
accounting for 200 noun pairs, 150 verb pairs and 50 adjective
pairs. Regarding noun pairs, we balanced the size of pairs in terms of
six relations: the five extracted relations from VCL and VWN, and an
``unrelated'' relation. For verb pairs, we balanced the number of
pairs according to five relations: synonymy, antonymy, hypernymy,
co-hyponymy, and unrelated. For adjective pairs, we balanced the size
of pairs for three relations: synonymy, antonymy, and unrelated. In
order to select the unrelated pairs for each part-of-speech category,
we paired the unrelated words from the selected related pairs at
random. From these random pairs, we excluded those pairs that appeared
in VCL and VWN. Furthermore, we also balanced the number of selected
pairs according to the sizes of the morphological classes and the
lexical categories.

\subsection{Annotation of ViSim-400}

For rating ViSim-400, 200 raters who were native Vietnamese speakers
were paid to rate the degrees of similarity for all 400 pairs. Each
rater was asked to rate 30 pairs on a 0--6 scale; and each pair was
rated by 15 raters. Unlike other datasets which performed the
annotation via Amazon Mechanical Turk, each rater for ViSim-400
conducted the annotation via a survey which detailed the exact
annotation guidelines.

The structure of the questionnaire was motivated by the SimLex-999
dataset: we outlined the notion of similarity via the well-understood
idea of the six relations included in the ViSim-400
dataset. Immediately after the guidelines of the questionnaire, a
checkpoint question was posed to the participants to test whether the
person understood the guidelines: the participant was asked to pick
the most similar word pair from three given word pairs, such as
\VI{kiêu\_căng/kiêu\_ngạo} (\textit{arrogant/cocky}) vs. \VI{trầm/bổng}
(\textit{high/low}) vs. \VI{cổ\_điển/biếng}
(\textit{classical/lazy}). The annotators then labeled the kind of
relation and scored the degree of similarity for each word pair in the
survey.

\subsection{Agreement in ViSim-400}

\begin{figure*}[t]
  \centering
  \resizebox{0.85\linewidth}{!}{%
    \includegraphics[draft=false,trim=1.cm 1.5cm 0cm 0cm 0,width=0.99\linewidth]{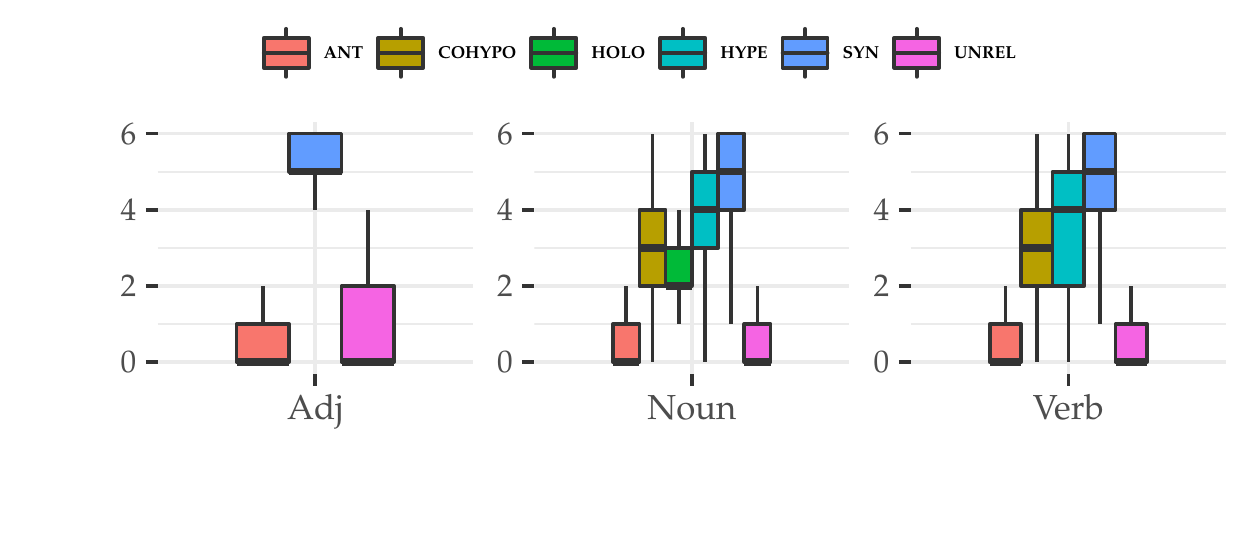}
  }
  \vspace{-2mm}
  \caption{Distribution of scored pairs in ViSim-400 across parts-of-speech and semantic relations.}
  \label{fig:plots}
  \vspace{+3mm}
\end{figure*}%

We analysed the ratings of the ViSim-400 annotators with two different
inter-annotator agreement (IAA) measures, Krippendorff's alpha
coefficient \cite{Krippendorff:04}, and the average standard deviation
(STD) of all pairs across word classes. The first IAA measure,
IAA-pairwise, computes the average pairwise Spearman’s $\rho$
correlation between any two raters. This IAA measure has been a common
choice in previous data collections in distributional semantics
\cite{PadoEtAl:2007,Reisinger/Mooney:2010,Hill:2015}. The second IAA
measure, IAA-mean, compares the average correlation of the human
raters with the average of all other raters. This measure would smooth
individual annotator effects, and serve as a more appropriate
\textit{``upper bound"} for the performance of automatic systems than
IAA-pairwise \cite{Vulic:16}. Finally, Krippendorff's $\alpha$
coefficient reflects the disagreement of annotators rather than their
agreement, in addition to correcting for agreement by chance.

Table~\ref{tbl:iaa} shows the inter-annotator agreement values,
Krippendorff's $\alpha$ coefficient, and the response consistency
measured by STD over all pairs and different word classes in
ViSim-400. The overall IAA-pairwise of ViSim-400 is $\rho=0.79$,
comparing favourably with the agreement on the SimLex-999 dataset
($\rho=0.67$ using the same IAA-pairwise measure). Regarding IAA-mean,
ViSim-400 also achieves an overall agreement of $\rho=0.86$, which is
similar to the agreement in \newcite{Vulic:16}, $\rho=0.86$. For
Krippendorff's $\alpha$ coefficient, the value achieves $\alpha =
0.78$, also reflecting the reliability of the annotated dataset.

\begin{table}[]
  \centering
  \resizebox{0.95\linewidth}{!}{%
    \begin{tabular}{@{}lcccc@{}}
      \toprule
      & \textbf{\textit{All}} & \textbf{Noun} & \textbf{Verb} & \textbf{Adjective} \\ \midrule
      IAA-Mean $\rho$ & 0.86         & 0.86          & 0.86          & 0.78               \\
      IAA-Pairwise $\rho$    & 0.79         & 0.76          & 0.78          & 0.75               \\
      Krippendorff's $\alpha$	& 0.78		& 0.76		& 0.78		& 0.86		\\
      STD	& 0.87	& 0.87	& 0.90	& 0.82		\\
      \bottomrule
    \end{tabular}
  }
  \caption{Inter-annotator agreements measured by Spearman’s $\rho$,
    Krippendorff's $\alpha$, and the average standard deviation (STD)
    of all pairs across word classes.}
  \label{tbl:iaa}
  \vspace{-2mm}
\end{table}

Furthermore, the box plots in Figure~\ref{fig:plots} present the
distributions of all rated pairs in terms of the fine-grained semantic
relations across word classes. They reveal that --across word
classes-- synonym pairs are clearly rated as the most similar words,
and antonym as well as unrelated pairs are clearly rated as the most
dissimilar words. Hypernymy, co-hyponymy and holonymy are in between,
but rather similar than dissimilar.


\section{Verification of Datasets}
\label{evaluating-models}

In this section, we verify our novel datasets ViCon and ViSim-400
through standard and neural co-occurrence models, in order to show
that we obtain a similar behaviour as for the corresponding English
datasets.

\subsection{Verification of ViSim-400}
\label{evaluating-models-ViSim-400}

We adopt a comparison of neural models on SimLex-999 as suggested by
\newcite{NguyenEtAl:2016}. They applied three models, a Skip-gram
model with negative sampling SGNS~\cite{Mikolov:2013}, the dLCE
model~\cite{NguyenEtAl:2016}, and the mLCM
model~\cite{Nghia:2015}. Both the dLCE and the mLCM models integrated
lexical contrast information into the basic Skip-gram model to train
word embeddings for distinguishing antonyms from synonyms, and for
reflecting degrees of similarity.

The three models were trained with 300 dimensions, a window size of 5
words, and 10 negative samples. Regarding the corpora, we relied on
Vietnamese corpora with a total of $\approx$145 million tokens,
including the \textit{Vietnamese Wikipedia},\footnote{\scriptsize
  \url{https://dumps.wikimedia.org/viwiki/latest/}}
\textit{VNESEcorpus} and \textit{VNTQcorpus},\footnote{\scriptsize
  \url{http://viet.jnlp.org/download-du-lieu-tu-vung-corpus}} and the
\textit{Leipzig Corpora Collection for
  Vietnamese}\footnote{\scriptsize
  \url{http://wortschatz.uni-leipzig.de/en/download}}
\cite{Goldhahn:2012}. For word segmentation and POS tagging, we used
the open-source toolkit \textit{UETnlp}\footnote{\scriptsize
  \url{https://github.com/phongnt570/UETnlp}} \cite{NguyenTP:2016}.
The antonym and synonym pairs to train the dLCE and mLCM models were
extracted from VWN consisting of 49,458 antonymous pairs and 338,714
synonymous pairs. All pairs which appeared in ViSim-400 were excluded
from this set.

Table~\ref{tbl:correlation} shows Spearman's correlations $\rho$,
comparing the scores of the three models with the human judgements for
ViSim-400. As also reported for English, the dLCE model produces the
best performance, SGNS the worst.

\begin{table}[h]
  \centering
  \resizebox{0.7\linewidth}{!}{%
    \begin{tabular}{@{}lccc@{}}
      \toprule
      & \textbf{SGNS} & \textbf{mLCM} & \textbf{dLCE} \\ \midrule
      ViSim-400 & 0.37          & 0.60          & \textbf{0.62} \\ 
      SimLex-999 & 0.38		  & 0.51			  & \textbf{0.59} \\ \bottomrule
    \end{tabular}
  }
  \caption{Spearman's correlation $\rho$ on ViSim-400 in comparison to SimLex-999, cf. \newcite{NguyenEtAl:2016}.}
  \label{tbl:correlation}
\end{table}

In a second experiment, we computed the cosine similarities between
all word pairs, and used the area under curve (AUC) to distinguish
between antonyms and synonyms.  Table~\ref{tbl:auc} presents the AUC results
of the three models. Again, the models show a similar behaviour in
comparison to SimLex-999, where also the dLCE model outperforms the
two other models, and the SGNS model is by far the worst.

\begin{table}[h]
  \centering
  \resizebox{0.8\linewidth}{!}{%
    \begin{tabular}{@{}llccc@{}}
      \toprule
      & \textbf{Model} & \textbf{Noun} & \textbf{Verb} & \textbf{Adj}  \\ \midrule
      \multirow{3}{*}{ViSim-400}  & SGNS           & 0.66          & 0.63          & 0.70          \\
      & mLCM           & 0.81          & 0.92          & 0.96          \\
      & dLCE           & \textbf{0.92} & \textbf{0.95} & \textbf{0.98} \\ \midrule
      \multirow{3}{*}{SimLex-999} & SGNS           & 0.66          & 0.65          & 0.64          \\
      & mLCM           & 0.69          & 0.71          & 0.85          \\
      & dLCE           & \textbf{0.72} & \textbf{0.81} & \textbf{0.90} \\ 
      \bottomrule
    \end{tabular}
  }
  \caption{AUC scores for distinguishing antonyms from synonyms in ViSim-400.}
  \label{tbl:auc}
\end{table}
\vspace{-1em}

\subsection{Verification of ViCon}
\label{evaluating-models-ViCon}

In order to verify ViCon, we applied three co-occurrence models to
rank antonymous and synonymous word pairs according to their cosine
similarities: two standard co-occurrence models based on positive
point-wise mutual information (PPMI) and positive local mutual
information (PLMI) \cite{Evert:2005} as well as an improved feature
value representation $\mathbf{weight^{SA}}$ as suggested by
\newcite{NguyenEtAl:2016}. For building the vector space co-occurrence
models, we relied on the same Vietnamese corpora as in the previous
section. For inducing the word vector representations via
$\mathbf{weight^{SA}}$, we made use of the antonymous and synonymous
pairs in VWN, as in the previous section, and then removed all pairs
which appeared in ViCon. Optionally, we applied singular value
decomposition (SVD) to reduce the dimensionalities of the word vector
representations.

As in \newcite{NguyenEtAl:2016}, we computed the cosine similarities
between all word pairs, and then sorted the pairs according to their
cosine scores. Average Precision (AP) evaluated the three vector space
models. Table~\ref{tbl:co-occurrence-models} presents the results of
the three vector space models with and without SVD. As for English,
the results on the Vietnamese dataset demonstrate significant
improvements ($\chi^2, ^* p<.001$) of $\mathbf{weight^{SA}}$ over PPMI
and PLMI, both with and without SVD, and across word classes.

\begin{table}[h]
\centering
\resizebox{0.99\linewidth}{!}{%
\begin{tabular}{clcccccc}
\hline
\multicolumn{2}{c}{}                                                            & \multicolumn{2}{c}{\textbf{ADJ}}                                                & \multicolumn{2}{c}{\textbf{NOUN}}                                               & \multicolumn{2}{c}{\textbf{VERB}}                                               \\ \cline{3-8} 
\multicolumn{2}{c}{\multirow{-2}{*}{\textbf{Metric}}}                           & SYN                                    & ANT                                    & SYN                                    & ANT                                    & SYN                                    & ANT                                    \\ \hline
                         & \cellcolor[HTML]{D99696}PPMI                         & \cellcolor[HTML]{D99696}0.70           & \cellcolor[HTML]{D99696}0.38           & \cellcolor[HTML]{D99696}0.68           & \cellcolor[HTML]{D99696}0.39           & \cellcolor[HTML]{D99696}0.69           & \cellcolor[HTML]{D99696}0.38           \\
                         & \cellcolor[HTML]{D99696}PLMI                         & \cellcolor[HTML]{D99696}0.59           & \cellcolor[HTML]{D99696}0.44           & \cellcolor[HTML]{D99696}0.61           & \cellcolor[HTML]{D99696}0.42           & \cellcolor[HTML]{D99696}0.63           & \cellcolor[HTML]{D99696}0.41           \\
                         & \cellcolor[HTML]{D99696}$\mathbf{weight^{SA}}$       & \cellcolor[HTML]{D99696}\textbf{0.93*} & \cellcolor[HTML]{D99696}\textbf{0.31*} & \cellcolor[HTML]{D99696}\textbf{0.94*} & \cellcolor[HTML]{D99696}\textbf{0.31}  & \cellcolor[HTML]{D99696}\textbf{0.96}  & \cellcolor[HTML]{D99696}\textbf{0.31}  \\
                         & \cellcolor[HTML]{34CDF9}PPMI + SVD                   & \cellcolor[HTML]{34CDF9}0.76           & \cellcolor[HTML]{34CDF9}0.36           & \cellcolor[HTML]{34CDF9}0.66           & \cellcolor[HTML]{34CDF9}0.40           & \cellcolor[HTML]{34CDF9}0.81           & \cellcolor[HTML]{34CDF9}0.34           \\
                         & \cellcolor[HTML]{34CDF9}PLMI + SVD                   & \cellcolor[HTML]{34CDF9}0.49           & \cellcolor[HTML]{34CDF9}0.51           & \cellcolor[HTML]{34CDF9}0.55           & \cellcolor[HTML]{34CDF9}0.46           & \cellcolor[HTML]{34CDF9}0.51           & \cellcolor[HTML]{34CDF9}0.49           \\
\multirow{-6}{*}{ViCon}  & \cellcolor[HTML]{34CDF9}$\mathbf{weight^{SA}}$ + SVD & \cellcolor[HTML]{34CDF9}\textbf{0.91*} & \cellcolor[HTML]{34CDF9}\textbf{0.32*} & \cellcolor[HTML]{34CDF9}\textbf{0.81*} & \cellcolor[HTML]{34CDF9}\textbf{0.34*} & \cellcolor[HTML]{34CDF9}\textbf{0.92*} & \cellcolor[HTML]{34CDF9}\textbf{0.32*} \\ \hline
                         & \cellcolor[HTML]{D99696}PLMI                         & \cellcolor[HTML]{D99696}0.56           & \cellcolor[HTML]{D99696}0.46           & \cellcolor[HTML]{D99696}0.60           & \cellcolor[HTML]{D99696}0.42           & \cellcolor[HTML]{D99696}0.62           & \cellcolor[HTML]{D99696}0.42           \\
                         & \cellcolor[HTML]{D99696}$\mathbf{weight^{SA}}$       & \cellcolor[HTML]{D99696}\textbf{0.75}  & \cellcolor[HTML]{D99696}\textbf{0.36}  & \cellcolor[HTML]{D99696}\textbf{0.66}  & \cellcolor[HTML]{D99696}\textbf{0.40}  & \cellcolor[HTML]{D99696}\textbf{0.71}  & \cellcolor[HTML]{D99696}\textbf{0.38}  \\
                         & \cellcolor[HTML]{34CDF9}PLMI + SVD                   & \cellcolor[HTML]{34CDF9}0.55           & \cellcolor[HTML]{34CDF9}0.46           & \cellcolor[HTML]{34CDF9}0.55           & \cellcolor[HTML]{34CDF9}0.46           & \cellcolor[HTML]{34CDF9}0.58           & \cellcolor[HTML]{34CDF9}0.44           \\
\multirow{-4}{*}{LexCon} & \cellcolor[HTML]{34CDF9}$\mathbf{weight^{SA}}$ + SVD & \cellcolor[HTML]{34CDF9}\textbf{0.76*} & \cellcolor[HTML]{34CDF9}\textbf{0.36*} & \cellcolor[HTML]{34CDF9}\textbf{0.66}  & \cellcolor[HTML]{34CDF9}\textbf{0.40}  & \cellcolor[HTML]{34CDF9}\textbf{0.70*} & \cellcolor[HTML]{34CDF9}\textbf{0.38*} \\ \hline
\end{tabular}
}
\caption{AP evaluation of co-occurrence models on ViCon in comparison to LexCon \cite{NguyenEtAl:2016}.}
\label{tbl:co-occurrence-models}
\end{table}

\section{Conclusion}

This paper introduced two novel datasets for the low-resource language
Vietnamese to assess models of semantic similarity: \textit{ViCon}
comprises synonym and antonym pairs across the word classes of
nouns, verbs, and adjectives. It offers data to distinguish between
similarity and dissimilarity. \textit{ViSim-400} contains 400 word
pairs across the three word classes and five semantic relations. Each
pair was rated by human judges for its degree of similarity, to
reflect the continuum between similarity and relatedness. The two
datasets were verified through standard co-occurrence and neural
network models, showing results comparable to the respective English
datasets.

\vspace{+2mm}
\section*{Acknowledgments}

The research was supported by the Ministry of Education and Training
of the Socialist Republic of Vietnam (Scholarship 977/QD-BGDDT;
Kim-Anh Nguyen), and the DFG Collaborative Research Centre SFB 732
(Kim-Anh Nguyen, Sabine Schulte im Walde, Ngoc Thang Vu).

\bibliography{naaclhlt2018}

\begin{thebibliography}{}
\expandafter\ifx\csname natexlab\endcsname\relax\def\natexlab#1{#1}\fi

\bibitem[{Biemann(2005)}]{Biemann:05}
Chris Biemann. 2005.
\newblock Ontology learning from text: A survey of methods.
\newblock {\em LDV Forum\/} 20(2):75--93.

\bibitem[{Bruni et~al.(2012)Bruni, Boleda, Baroni, and Tran}]{Bruni:2012}
Elia Bruni, Gemma Boleda, Marco Baroni, and Nam-Khanh Tran. 2012.
\newblock Distributional semantics in technicolor.
\newblock In {\em Proceedings of the 50th Annual Meeting of the Association for
  Computational Linguistics\/}. Jeju Island, Korea, pages 136--145.

\bibitem[{Budanitsky and Hirst(2006)}]{Budanitsky/Hirst:06}
Alexander Budanitsky and Graeme Hirst. 2006.
\newblock Evaluating {W}ord{N}et-based measures of lexical semantic
  relatedness.
\newblock {\em Computational Linguistics\/} 32(1):13--47.

\bibitem[{Cimiano et~al.(2005)Cimiano, Hotho, and Staab}]{Cimiano:05}
Philipp Cimiano, Andreas Hotho, and Steffen Staab. 2005.
\newblock Learning concept hierarchies from text corpora using formal concept
  analysis.
\newblock {\em Journal of Artificial Intelligence Research\/} 24(1):305--339.

\bibitem[{Evert(2005)}]{Evert:2005}
Stefan Evert. 2005.
\newblock {\em The Statistics of Word Cooccurrences\/}.
\newblock Ph.D. thesis, University of Stuttgart.

\bibitem[{Finkelstein et~al.(2001)Finkelstein, Gabrilovich, Matias, Rivlin,
  Solan, Wolfman, and Ruppin}]{Finkelstein:2001}
Lev Finkelstein, Evgeniy Gabrilovich, Yossi Matias, Ehud Rivlin, Zach Solan,
  Gadi Wolfman, and Eytan Ruppin. 2001.
\newblock Placing search in context: The concept revisited.
\newblock In {\em Proceedings of the 10th International Conference on World
  Wide Web\/}. Hong Kong, Hong Kong, pages 406--414.

\bibitem[{Gentner(2006)}]{Gentner:06}
Dedre Gentner. 2006.
\newblock Why verbs are hard to learn.
\newblock In Kathryn~A. Hirsh-Pasek and Roberta~M. Golinkoff, editors, {\em
  Action meets word: How Children Learn Verbs\/}, Oxford University Press,
  pages 544--564.

\bibitem[{Goldhahn et~al.(2012)Goldhahn, Eckart, and Quasthoff}]{Goldhahn:2012}
Dirk Goldhahn, Thomas Eckart, and Uwe Quasthoff. 2012.
\newblock Building large monolingual dictionaries at the {L}eipzig corpora
  collection: From 100 to 200 languages.
\newblock In {\em Proceedings of the 8th International Conference on Language
  Resources and Evaluation\/}. pages 759--765.

\bibitem[{Gurevych(2005)}]{Gurevych:2005}
Iryna Gurevych. 2005.
\newblock Using the structure of a conceptual network in computing semantic
  relatedness.
\newblock In {\em Proceedings of the 2nd International Joint Conference on
  Natural Language Processing\/}. Jeju Island, Republic of Korea, pages
  767--778.

\bibitem[{Gurevych(2006)}]{Gurevych:2006}
Iryna Gurevych. 2006.
\newblock Thinking beyond the nouns: Computing semantic relatedness across
  parts of speech.
\newblock In {\em Proceedings of Sprachdokumentation \& Sprachbeschreibung, 28.
  Jahrestagung der Deutschen Gesellschaft f{\"u}r Sprachwissenschaft\/}.
  Bielefeld, Germany, page 226.

\bibitem[{He et~al.(2008)He, Yang, Gao, Nguyen, and Moore}]{He:08}
Xiaodong He, Mei Yang, Jianfeng Gao, Patrick Nguyen, and Robert Moore. 2008.
\newblock Indirect-{HMM}-based hypothesis alignment for combining outputs from
  machine translation systems.
\newblock In {\em Proceedings of the Conference on Empirical Methods in Natural
  Language Processing\/}. Honolulu, Hawaii, pages 98--107.

\bibitem[{Hill et~al.(2015)Hill, Reichart, and Korhonen}]{Hill:2015}
Felix Hill, Roi Reichart, and Anna Korhonen. 2015.
\newblock Simlex-999: Evaluating semantic models with genuine similarity
  estimation.
\newblock {\em Computational Linguistic\/} 41(4):665--695.

\bibitem[{Krippendorff(2004)}]{Krippendorff:04}
Klaus Krippendorff. 2004.
\newblock {\em Content Analysis: An Introduction to its Methodology\/}.
\newblock Sage Publications.

\bibitem[{Landauer and Dumais(1997)}]{Landauer/Dutnais:1997}
Thomas~K. Landauer and Susan~T. Dumais. 1997.
\newblock A solution to {P}lato's problem: The latent semantic analysis theory
  of acquisition, induction and representation of knowledge.
\newblock {\em Psychological Review\/} 104(2):211--240.

\bibitem[{Leviant and Reichart(2015)}]{Leviant/Reichart:2015}
Ira Leviant and Roi Reichart. 2015.
\newblock Judgment language matters: Multilingual vector space models for
  judgment language aware lexical semantics.
\newblock {\em CoRR\/} abs/1508.00106.

\bibitem[{Li et~al.(2006)Li, Zhang, Zhu, and Zhou}]{Li:06}
Mu~Li, Yang Zhang, Muhua Zhu, and Ming Zhou. 2006.
\newblock Exploring distributional similarity-based models for query spelling
  correction.
\newblock In {\em Proceedings of the 44th Annual Meeting of the Association for
  Computational Linguistics\/}. Sydney, Australia, pages 1025--1032.

\bibitem[{Marton et~al.(2009)Marton, Callison-Burch, and Resnik}]{Marton:09}
Yuval Marton, Chris Callison-Burch, and Philip Resnik. 2009.
\newblock Improved statistical machine translation using monolingually-derived
  paraphrases.
\newblock In {\em Proceedings of the 2009 Conference on Empirical Methods in
  Natural Language Processing\/}. Singapore, pages 381--390.

\bibitem[{Mikolov et~al.(2013)Mikolov, Sutskever, Chen, Corrado, and
  Dean}]{Mikolov:2013}
Tomas Mikolov, Ilya Sutskever, Kai Chen, Greg Corrado, and Jeffrey Dean. 2013.
\newblock Distributed representations of words and phrases and their
  compositionality.
\newblock In {\em Proceedings of the 26th International Conference on Neural
  Information Processing Systems\/}. Lake Tahoe, Nevada, pages 3111--3119.

\bibitem[{Morris and Hirst(2004)}]{Morris/Hirst:2004}
Jane Morris and Graeme Hirst. 2004.
\newblock Non-classical lexical semantic relations.
\newblock In {\em Proceedings of the HLT-NAACL Workshop on Computational
  Lexical Semantics\/}. Boston, Massachusetts, pages 46--51.

\bibitem[{Nguyen et~al.(2017)Nguyen, {Schulte im Walde}, and
  Vu}]{NguyenEtAl:2017}
Kim~Anh Nguyen, Sabine {Schulte im Walde}, and Ngoc~Thang Vu. 2017.
\newblock Distinguishing antonyms and synonyms in a pattern-based neural
  network.
\newblock In {\em Proceedings of the 15th Conference of the European Chapter of
  the Association for Computational Linguistics\/}. Valencia, Spain, pages
  76--85.

\bibitem[{Nguyen et~al.(2016{\natexlab{a}})Nguyen, {Schulte im Walde}, and
  Vu}]{NguyenEtAl:2016}
Kim-Anh Nguyen, Sabine {Schulte im Walde}, and Thang Vu. 2016{\natexlab{a}}.
\newblock Integrating distributional lexical contrast into word embeddings for
  antonym-synonym distinction.
\newblock In {\em Proceedings of the 54th Annual Meeting of the Association for
  Computational Linguistics\/}. Berlin, Germany, pages 454--459.

\bibitem[{Nguyen et~al.(2016{\natexlab{b}})Nguyen, Pham, Nguyen, Vu, Tran, and
  Truong}]{NguyenPT:2016}
Phuong-Thai Nguyen, Van-Lam Pham, Hoang-An Nguyen, Huy-Hien Vu, Ngoc-Anh Tran,
  and Thi-Thu-Ha Truong. 2016{\natexlab{b}}.
\newblock A two-phase approach for building a {V}ietnamese {W}ord{N}et.
\newblock In {\em Proceedings of the 8th Global WordNet Conference\/}.
  Bucharest, Romania, pages 259--264.

\bibitem[{Nguyen et~al.(2006)Nguyen, Romary, Rossignol, and
  Vu}]{NguyenTMH:2006}
Thi Minh~Huyen Nguyen, Laurent Romary, Mathias Rossignol, and Xuan~Luong Vu.
  2006.
\newblock {A lexicon for Vietnamese language processing}.
\newblock {\em {Language Resources and Evaluation}\/} 40(3-4):291--309.

\bibitem[{Nguyen and Le(2016)}]{NguyenTP:2016}
Tuan-Phong Nguyen and Anh-Cuong Le. 2016.
\newblock A hybrid approach to {V}ietnamese word segmentation.
\newblock In {\em Proceedings of the International Conference on Computing
  Communication Technologies, Research, Innovation, and Vision for the
  Future\/}. Hanoi, Vietnam, pages 114--119.

\bibitem[{Pad{ó} et~al.(2007)Pad{ó}, Pad{ó}, and Erk}]{PadoEtAl:2007}
Sebastian Pad{ó}, Ulrike Pad{ó}, and Katrin Erk. 2007.
\newblock Flexible, corpus-based modelling of human plausibility judgements.
\newblock In {\em Proceedings of the joint Conference on Empirical Methods in
  Natural Language Processing and Computational Natural Language Learning\/}.
  Prague, Czech Republic.

\bibitem[{Pham et~al.(2015)Pham, Lazaridou, and Baroni}]{Nghia:2015}
Nghia~The Pham, Angeliki Lazaridou, and Marco Baroni. 2015.
\newblock A multitask objective to inject lexical contrast into distributional
  semantics.
\newblock In {\em Proceedings of the 53rd Annual Meeting of the Association for
  Computational Linguistics and the 7th International Joint Conference on
  Natural Language Processing\/}. Beijing, China, pages 21--26.

\bibitem[{Reisinger and Mooney(2010)}]{Reisinger/Mooney:2010}
Joseph Reisinger and Raymond Mooney. 2010.
\newblock A mixture model with sharing for lexical semantics.
\newblock In {\em Proceedings of the 2010 Conference on Empirical Methods in
  Natural Language Processing\/}. Cambridge, Massachusetts, pages 1173--1182.

\bibitem[{Rubenstein and Goodenough(1965)}]{Rubenstein/Goodenough:1965}
Herbert Rubenstein and John~B. Goodenough. 1965.
\newblock Contextual correlates of synonymy.
\newblock {\em Communications of the ACM\/} 8(10):627--633.

\bibitem[{Scheible and {Schulte im Walde}(2014)}]{Scheible/Schulteimwalde:2014}
Silke Scheible and Sabine {Schulte im Walde}. 2014.
\newblock A database of paradigmatic semantic relation pairs for {G}erman
  nouns, verbs and adjectives.
\newblock In {\em Proceedings of the COLING Workshop \textit{Lexical and
  Grammatical Resources for Language Processing}\/}. Dublin, Ireland, pages
  111--119.

\bibitem[{Vuli{\'{c}} et~al.(2017)Vuli{\'{c}}, Gerz, Kiela, Hill, and
  Korhonen}]{Vulic:16}
Ivan Vuli{\'{c}}, Daniela Gerz, Douwe Kiela, Felix Hill, and Anna Korhonen.
  2017.
\newblock Hyperlex: A large-scale evaluation of graded lexical entailment.
\newblock {\em Computational Linguistics\/} 43(4):781--835.

\end{thebibliography}
\bibliographystyle{acl_natbib}

\end{document}